\documentclass[12pt]{article}

\usepackage[margin=1in]{geometry}

\usepackage{amsmath,amssymb,amsfonts}
\usepackage{graphicx}
\usepackage{booktabs}
\usepackage{multirow}
\usepackage{hyperref}
\usepackage{url}
\usepackage{xcolor}
\usepackage{microtype}
\usepackage[numbers,sort&compress]{natbib}

\hypersetup{
  colorlinks = true,
  linkcolor  = blue,
  citecolor  = blue,
  urlcolor   = blue
}

\title{Quantization Impact on the Accuracy and Communication Efficiency\\
Trade-off in Federated Learning for Aerospace Predictive Maintenance}

\author{Abdelkarim LOUKILI\\
\texttt{abdelkarim.loukili@ens-paris-saclay.fr}}

\date{\today}

\begin{document}

\maketitle

\begin{abstract}
Federated learning (FL) enables privacy-preserving predictive maintenance
across distributed aerospace fleets, but gradient communication overhead
constrains deployment on bandwidth-limited IoT nodes.
This paper investigates the impact of symmetric uniform quantization
($b \in \{32,8,4,2\}$ bits) on the accuracy--efficiency trade-off of a
custom-designed lightweight 1-D convolutional model (AeroConv1D,
9\,697 parameters) trained via FL on the NASA C-MAPSS benchmark
under a realistic Non-IID client partition.
Using a rigorous multi-seed evaluation ($N=10$ seeds), we show that
INT4 achieves accuracy \emph{statistically indistinguishable} from
FP32 on both FD001 ($p=0.341$) and FD002 ($p=0.264$ MAE,
$p=0.534$ NASA score) while delivering an $8\times$ reduction in
gradient communication cost (37.88~KiB $\to$ 4.73~KiB per round).
A key methodological finding is that na\"ive IID client partitioning
artificially suppresses variance; correct Non-IID evaluation reveals
the true operational instability of extreme quantization, demonstrated
via a direct empirical IID vs.\ Non-IID comparison.
INT2 is empirically characterized as unsuitable: while it
achieves lower MAE on FD002 through extreme quantization-induced
over-regularization, this apparent gain is accompanied by catastrophic
NASA score instability (CV\,=\,45.8\% vs.\ 22.3\% for FP32),
confirming non-reproducibility under heterogeneous operating conditions.
Analytical FPGA resource projections on the Xilinx ZCU102 confirm that
INT4 fits within hardware constraints (85.5\% DSP utilization),
potentially enabling a complete FL pipeline on a single SoC.
The full simulation codebase and FPGA estimation scripts are publicly
available at \url{https://github.com/therealdeadbeef/aerospace-fl-quantization}.
\end{abstract}

\section{Introduction}
\label{sec:intro}

Predictive maintenance of aerospace propulsion systems relies on
accurate estimation of the Remaining Useful Life (RUL) of turbofan
engines \citep{saxena2008}.
As aerospace operators increasingly operate large, geographically
distributed fleets, a fundamental tension arises: training accurate
predictive models requires pooling data across many engines, yet
centralizing raw telemetry raises significant privacy, regulatory,
and bandwidth concerns.
Federated learning (FL) \citep{mcmahan2017} resolves this tension by
training models collaboratively across edge nodes without exposing raw
data to a central server.

However, FL deployment in aerospace IoT settings faces two
compounding practical constraints.
First, \emph{communication overhead}: each FL round requires
broadcasting a full-precision gradient vector, whose size scales
linearly with model precision. Over bandwidth-constrained aeronautical
links (e.g., LoRaWAN at 5~kbps), even modest models become
prohibitively expensive to synchronize.
Second, \emph{hardware constraints}: inference must run on
resource-constrained FPGAs rather than cloud GPUs, imposing
strict limits on model complexity and numerical precision.

Symmetric uniform gradient quantization addresses both constraints
simultaneously by reducing the bit-width $b$ of transmitted gradients.
Lower-precision updates occupy fewer bits per parameter, directly
reducing communication cost; lower-precision arithmetic also reduces
FPGA resource utilization, enabling deployment on smaller devices.
However, the quantization--accuracy trade-off in FL has been studied
almost exclusively under IID data assumptions and on general-purpose
classification benchmarks \citep{alistarh2017,bernstein2018,ma2022},
leaving open questions about its behavior under the Non-IID
distributions that characterize realistic aerospace deployments with
heterogeneous operating conditions \citep{purkayastha2024}.

This paper makes four contributions:

\begin{enumerate}

\item \textbf{AeroConv1D and experimental protocol.}
We design AeroConv1D, a custom sub-10k parameter, purely feed-forward
1-D CNN optimized for FPGA inference, and conduct a multi-seed
($N=10$), Non-IID evaluation of four quantization levels
($b \in \{32,8,4,2\}$) on NASA C-MAPSS FD001 and FD002 using paired
$t$-tests to assess statistical significance.

\item \textbf{Methodological contribution on IID bias.}
We demonstrate empirically that IID-biased client partitioning
artificially suppresses variance and inflates the apparent accuracy
benefit of quantization. Under correct Non-IID evaluation, INT4
achieves \emph{accuracy parity} with FP32 ($p>0.05$ on all
metric $\times$ subset combinations) rather than dominance.

\item \textbf{Characterization of INT2 instability.}
We show that INT2 exhibits an unexpected MAE reduction on FD002
attributed to extreme over-regularization by the 3-level quantization
grid, accompanied by catastrophic NASA score instability
(CV\,=\,45.8\%), making it operationally unusable regardless of its
average error.

\item \textbf{Hardware projection.}
Analytical FPGA resource projections following the hls4ml scaling
model \citep{hls4ml2021} show that INT4 fits within the Xilinx ZCU102
(85.5\% DSP), leaving 366 spare DSPs for a potential NTT-based
homomorphic encryption co-processor \citep{nguyen2023}.

\end{enumerate}

\paragraph{Scope and limitations.}
While previous iterations of this work emphasized a gradient-distortion
privacy proxy, we recognize this metric serves only as a heuristic
indicator of gradient-inversion attack surface
\citep{zhu2019,geiping2020} rather than a formal $(\varepsilon,\delta)$-DP
bound; establishing formal DP guarantees for the RUL regression setting
is left as future work.
FPGA projections are analytical and have not been validated on physical
ZCU102 silicon; silicon validation is part of ongoing independent
research \citep{khalil2023,wang2023}.

\section{Related Work}
\label{sec:related}

\subsection{Federated Learning for Predictive Maintenance}

\citet{mcmahan2017} introduced FedAvg as a communication-efficient,
privacy-preserving distributed training paradigm.
\citet{landau2025} propose FL across multi-airline fleets for
RUL prediction, and \citet{purkayastha2024} survey FL's role in industrial
maintenance more broadly.
However, neither work addresses communication overhead or quantization
trade-offs on constrained edge nodes, the central concerns of the
present paper.

\subsection{Quantization in Federated Learning}

Quantization of gradients for communication efficiency has a
substantial literature.
\citet{alistarh2017} introduce QSGD, providing unbiased stochastic
quantization with convergence guarantees.
\citet{bernstein2018} propose SignSGD, which transmits only the sign
of each gradient component, achieving extreme compression at the cost
of bias.
\citet{ma2022} survey the broader challenges of resolving Non-IID data
distributions in FL. Concurrently, recent work by \citet{he2025} reports
that low-bit quantization can act as an implicit regularizer under certain
conditions, though they note results are dataset-dependent.

Recent advances have also explored hybrid and runtime quantization strategies.
\citet{zheng2025fedhqhybridruntimequantization} introduce FedHQ, a framework that dynamically
combines post-training and quantization-aware training at runtime to
automatically allocate optimal hybrid strategies per client under
heterogeneous FL conditions, further demonstrating the potential of
adaptive quantization as an implicit regularizer.

Our work revisits these claims on an aerospace RUL regression benchmark.
We find that the regularization effect is statistically indiscernible
for INT4 on both subsets after correcting for IID partitioning bias,
while INT2 produces a spurious MAE improvement on the harder FD002
subset that is operationally meaningless due to catastrophic score
instability.
This constitutes a methodological warning for practitioners who evaluate
quantization under IID assumptions and then deploy under Non-IID
conditions.

\paragraph{Scope of the quantization comparison.}
This work evaluates symmetric uniform per-tensor quantization as a
clean, hardware-deployable baseline rather than as an exhaustive survey
of FL compression schemes.
QSGD \citep{alistarh2017} adds stochastic rounding and variable-length
entropy coding, which reduces communication further but requires
floating-point dequantization at the aggregator and is not directly
implementable in fixed-point FPGA pipelines.
SignSGD \citep{bernstein2018} achieves 1-bit compression but
introduces gradient bias that can harm convergence under Non-IID
distributions \citep{ma2022}.
Advanced compression schemes \citep{he2025} dynamically assign bit-widths
or apply non-uniform mappings, which could improve the INT4 operating
point further but require complex decoding logic incompatible with the
strict latency budget of aeronautical IoT links.
Comparing these schemes head-to-head on C-MAPSS under Non-IID
conditions is a natural extension; we leave it to future work to avoid
conflating the methodological contribution (IID partitioning bias) with
a compression benchmark.

\subsection{FPGA Acceleration and Cryptographic Co-design}

hls4ml \citep{hls4ml2021} enables automatic synthesis of neural networks
to Xilinx FPGAs with configurable precision, providing the scaling model
we use for resource projections.
NTT-based homomorphic encryption (HE) accelerators have been demonstrated
on Zynq platforms \citep{nguyen2023,ye2025,laouiti2025}, motivating the
spare-DSP co-design goal of this work: if INT4 inference fits comfortably
on the ZCU102, the remaining DSP budget could host an HE co-processor,
enabling encrypted gradient transmission without a second device.

\section{System Model}
\label{sec:model}

\subsection{Federated Learning Framework}

We consider a synchronous FedAvg setup with $N=10$ clients and a
central aggregator.
Each client $k$ holds a private dataset $\mathcal{D}_k$ (a subset
of turbofan engine trajectories) and trains a local copy of the global
model for $E=2$ local epochs per round, producing updated weights
$\mathbf{w}^{r,(k)}$.
The aggregator updates the global model as:
\begin{equation}
  \mathbf{w}^{r+1} = \mathbf{w}^{r} +
    \frac{1}{N}\sum_{k=1}^{N}
    Q_b\!\left(\mathbf{w}^{r,(k)} - \mathbf{w}^{r}\right),
  \label{eq:fedavg}
\end{equation}
where $Q_b(\cdot)$ denotes symmetric uniform quantization to $b$~bits
applied to the per-client weight delta
$\Delta\mathbf{w}^{(k)} = \mathbf{w}^{r,(k)} - \mathbf{w}^{r}$
before transmission.
Quantization is applied to the delta, not to the model weights during
local training, preserving full-precision gradient accumulation on each
client.

\subsection{Local Model: AeroConv1D}
\label{sec:model_arch}

To meet the strict hardware constraints of aerospace IoT nodes, we
propose AeroConv1D, a custom lightweight 1-D convolutional architecture
(9\,697 parameters) designed specifically for this study.
Recurrent architectures (e.g., LSTMs) and CNN-LSTM hybrids are common
baselines for C-MAPSS RUL prediction, but their recurrent temporal
dependencies complicate ultra-low-bit quantization --- weight state
accumulation amplifies quantization noise across time steps --- and
prevent deep hardware pipelining, which is essential for low-latency
FPGA inference.
AeroConv1D instead relies on a purely feed-forward topology to maximize
FPGA parallelism.

The architecture processes temporal windows of 50 time steps over 14
variance-filtered sensor channels.
A small temporal kernel ($k=3$) efficiently captures local sensor
degradation trends without excessive multiplication overhead.
The subsequent channel doubling ($14\to32\to64$) builds a hierarchical
feature representation, providing sufficient capacity while strictly
bounding the total parameter footprint below 10k.
The full layer-by-layer specification is given in
Table~\ref{tab:architecture}.

\begin{table}[htbp]
\centering
\caption{AeroConv1D architecture specification (9,697 parameters total).
  $B$: batch size. The parameter count is verified programmatically at
  simulation startup via an assertion.}
\label{tab:architecture}
\begin{tabular}{clcc}
\toprule
\textbf{Layer} & \textbf{Type / Configuration} &
  \textbf{Output Shape} & \textbf{Params} \\
\midrule
Input & Time-series window & $(B, 14, 50)$ & 0 \\
1 & Conv1D ($k=3, s=1, p=1$) + ReLU & $(B, 32, 50)$ & 1,376 \\
2 & MaxPool1D ($k=2, s=2$) & $(B, 32, 25)$ & 0 \\
3 & Conv1D ($k=3, s=1, p=1$) + ReLU & $(B, 64, 25)$ & 6,208 \\
4 & AdaptiveAvgPool1D & $(B, 64, 1)$ & 0 \\
5 & Flatten & $(B, 64)$ & 0 \\
6 & Linear + ReLU & $(B, 32)$ & 2,080 \\
7 & Linear & $(B, 1)$ & 33 \\
\midrule
\multicolumn{3}{l}{\textbf{Total}} & \textbf{9,697} \\
\bottomrule
\end{tabular}
\end{table}

\subsection{Symmetric Uniform Quantization}
\label{sec:quant}

Prior to transmission, each client applies symmetric uniform
quantization independently to each layer's weight delta
$\Delta\mathbf{w}^{(l)}$:
\begin{equation}
  Q_b\!\left(\Delta\mathbf{w}^{(l)}\right) =
    \frac{\alpha^{(l)}}{2^{b-1}-1}
    \left\lfloor
      \frac{(2^{b-1}-1)\,\Delta\mathbf{w}^{(l)}}{\alpha^{(l)}}
    \right\rceil_{\rm clip},
  \label{eq:quant}
\end{equation}
where $\alpha^{(l)} = \max|\Delta\mathbf{w}^{(l)}|$ is the per-tensor
scale factor computed independently for each layer $l$, following
standard per-tensor quantization practice, and
$\lfloor\cdot\rceil_{\rm clip}$ denotes round-to-nearest followed by
saturation clipping to
$[-(2^{b-1}-1),\,+(2^{b-1}-1)]$.\footnote{The notation
$\lfloor\cdot\rceil_{\rm clip}$ is non-standard and introduced here
for compactness; it combines rounding (nearest integer) with symmetric
saturation clipping.}

For INT2, $2^{b-1}-1 = 1$, yielding the 3-level grid
$\{-\alpha^{(l)}, 0, +\alpha^{(l)}\}$.
This extreme coarseness is the root cause of INT2's over-regularization
behavior discussed in Section~\ref{sec:results_int2}.
Note that the INT2 grid $\{-\alpha^{(l)}, 0, +\alpha^{(l)}\}$ is
effectively a \emph{per-layer ternary update with a learned scale}
$\alpha^{(l)} = \max|\Delta\mathbf{w}^{(l)}|$, which differs from
the fixed $\{-1,0,+1\}$ grids used in ternary network classification
literature \citep{li2016}; the scale adapts each round to the
magnitude of the weight delta, so the scheme remains within the
symmetric uniform quantization family of Eq.~\eqref{eq:quant}
rather than constituting a separate ternarization algorithm.

\subsection{Dataset and Non-IID Client Partition}
\label{sec:nonIID}

The NASA C-MAPSS dataset \citep{saxena2008} provides run-to-failure
trajectories of turbofan engines under controlled degradation scenarios.
We use two subsets: FD001 (100 training engines, 1 operating condition)
and FD002 (260 training engines, 6 operating conditions).
RUL targets are capped at 125 cycles (piece-wise linear label).
The 14 variance-informative sensor channels retained are:
s2, s3, s4, s7, s8, s9, s11, s12, s13, s14, s15, s17, s20, s21.

Features are z-score standardized using training-set statistics, applied
identically to the test set.
Test-set ground-truth RUL values are loaded from the official
\texttt{RUL\_FDxxx.txt} files rather than inferred from cycle counts,
which would underestimate RUL for the truncated test sequences.
All available test windows (sliding over the full test trajectory,
approximately 8,700 windows for FD001 and 22,000 for FD002) are used
for evaluation, matching the NASA score formulation.

\paragraph{Non-IID partition.}
Client partitioning assigns engines per client, sampled
\emph{without replacement} and \emph{without sorting by RUL},
so that each client's RUL histogram differs from the global
distribution.
This corrects the IID-biased assignment common in preliminary
evaluations, which assigns contiguous engine blocks and artificially
homogenizes each client's data distribution.

To quantify the resulting heterogeneity, Table~\ref{tab:noniid} reports
the per-client mean RUL and Earth Mover's Distance (EMD) from the global
RUL distribution for seed~42.
The inter-client EMD spread (Avg.\ EMD\,=\,3.9 cycles on FD001,
2.8 cycles on FD002) confirms that the partition induces meaningful
label heterogeneity.

\begin{table}[htbp]
\centering
\caption{Per-client mean RUL (cycles) for seed~42, confirming Non-IID
  label heterogeneity. EMD: Earth Mover's Distance from the global
  RUL histogram.}
\label{tab:noniid}
\begin{tabular}{lcccc}
\toprule
 & \multicolumn{2}{c}{\textbf{FD001}} &
   \multicolumn{2}{c}{\textbf{FD002}} \\
\cmidrule(lr){2-3}\cmidrule(lr){4-5}
\textbf{Client} & Mean RUL & EMD & Mean RUL & EMD \\
\midrule
$k=1$  & 66.5 &  8.9 & 72.0 & 3.4 \\
$k=2$  & 74.1 &  1.2 & 77.6 & 2.2 \\
$k=3$  & 68.8 &  6.5 & 78.1 & 2.6 \\
$k=4$  & 73.3 &  2.0 & 74.8 & 0.6 \\
$k=5$  & 86.8 & 11.4 & 71.5 & 4.0 \\
$k=6$  & 74.8 &  0.5 & 72.2 & 3.2 \\
$k=7$  & 76.4 &  1.1 & 79.5 & 4.0 \\
$k=8$  & 76.8 &  1.4 & 79.7 & 4.2 \\
$k=9$  & 71.8 &  3.5 & 73.9 & 1.6 \\
$k=10$ & 77.7 &  2.4 & 73.4 & 2.0 \\
\midrule
Global   & 75.3 & ---  & 75.5 & --- \\
Avg.\ EMD & ---  & 3.9  & ---  & 2.8 \\
\bottomrule
\end{tabular}
\end{table}

\subsection{Evaluation Metrics}
\label{sec:metrics}

\paragraph{MAE.} Mean absolute error in RUL cycles:
$\text{MAE} = \frac{1}{n}\sum_i |\hat{y}_i - y_i|$.

\paragraph{NASA asymmetric score.}
\begin{equation}
  S = \sum_{i} s(d_i), \qquad
  s(d) = \begin{cases}
    e^{-d/13} - 1 & d < 0 \quad (\text{under-prediction})\\
    e^{d/10}  - 1 & d \geq 0 \quad (\text{over-prediction})
  \end{cases}
  \label{eq:nasa}
\end{equation}
where $d_i = \hat{y}_i - y_i$.
Over-prediction is penalised exponentially more steeply than
under-prediction, reflecting the safety-critical cost of declaring
a healthy engine as near-failure.
$S$ is reported as a sum over all test windows (approximately
8,700 for FD001; 22,000 for FD002), making it sensitive to both
systematic bias and prediction variance.

\begin{figure}[htbp]
  \centering
  \includegraphics[width=0.6\textwidth]{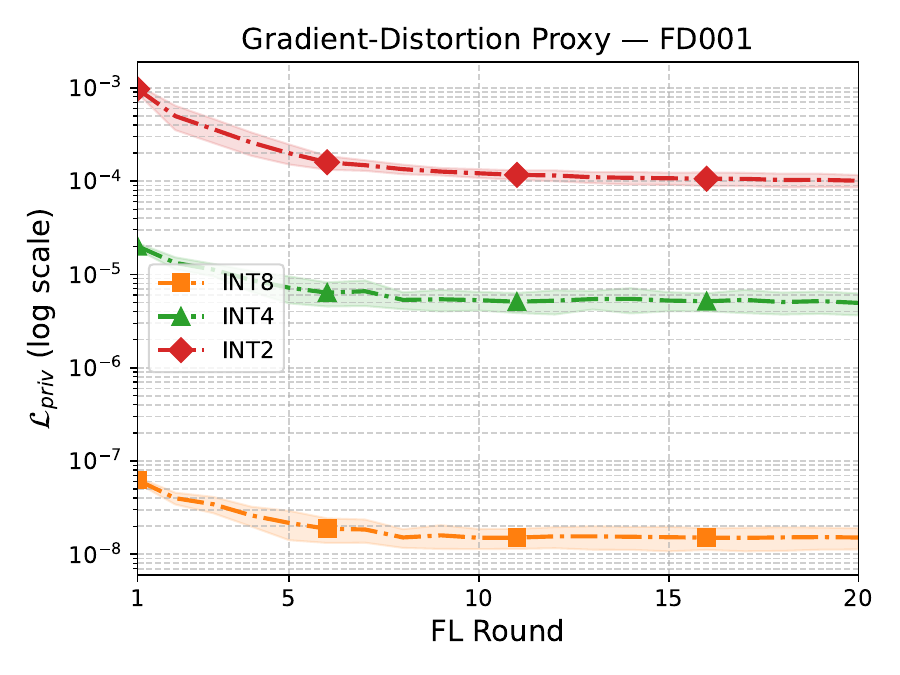}
  \caption{Gradient-distortion privacy proxy $\mathcal{L}_{\mathrm{priv}}$
    on FD001 (log scale) over 20 FL rounds.
    Higher values indicate greater gradient distortion and higher
    gradient-inversion attack cost \citep{zhu2019,geiping2020}.
    FP32 is omitted ($\mathcal{L}_{\mathrm{priv}}=0$ by definition).
    $\mathcal{L}_{\mathrm{priv}}$ is not a formal DP bound;
    see Section~\ref{sec:metrics}.}
  \label{fig:leakage}
\end{figure}

\paragraph{Gradient-distortion privacy proxy.}
\begin{equation}
  \mathcal{L}_{\mathrm{priv}} =
    \frac{1}{|\theta|}
    \bigl\|\Delta\mathbf{w} - Q_b(\Delta\mathbf{w})\bigr\|_2^2,
  \label{eq:leakage}
\end{equation}
where $|\theta| = 9{,}697$.
This measures the mean squared quantization distortion per parameter,
averaged over the $N$ clients per round.
Higher $\mathcal{L}_{\mathrm{priv}}$ indicates greater gradient
corruption, which raises the noise floor for gradient-inversion attacks
\citep{zhu2019,geiping2020}.
\emph{$\mathcal{L}_{\mathrm{priv}}$ is not a formal DP bound}; it is
used here solely as an exploratory indicator.
FP32 transmits the unquantized delta
($\mathcal{L}_{\mathrm{priv}}=0$ by definition) and is therefore
omitted from Figure~\ref{fig:leakage}.

\section{Experimental Setup}
\label{sec:setup}

\paragraph{Simulation protocol.}
Simulations run for 20 FL rounds with local batch size 32,
learning rate $10^{-3}$, and Adam optimiser.
To isolate the effect of distributional heterogeneity, a baseline IID
partition was additionally simulated on FD001.
The IID evaluation is restricted to FD001 for computational efficiency,
as it sufficiently demonstrates the baseline bias without requiring the
full FD002 parameter sweep.

\paragraph{Reproducibility.}
Each configuration is evaluated for $N=10$ random seeds
\texttt{\{42, 123, 256, 789, 1024, 2024, 3141, 4242, 5555, 9999\}},
controlling client partitioning, mini-batch shuffling, and weight
initialisation.
The local training RNG is set per-round and per-client as
$\text{seed}_{\text{client}} = s \cdot 10^4 + r \cdot 10^2 + k$,
ensuring statistically independent shuffles across rounds.

\paragraph{Statistical analysis.}
Results are reported as mean\,$\pm$\,std (sample std, $\text{df}=9$)
over seeds.
Statistical significance is assessed with a two-tailed paired
$t$-test ($\alpha=0.05$, $\text{df}=9$).
At $N=10$, the 95\% confidence interval on Cohen's $\hat{d}$ spans
approximately $\hat{d} \pm 0.95$ \citep{hedges1985}; effect-size
estimates are reported as directional indicators only.
The larger seed-to-seed variability observed under the corrected
Non-IID partition (e.g., FD001 FP32 NASA Score std\,=\,123k
vs.\ 41k under IID, see Table~\ref{tab:iid_bias}) further validates
the distributional heterogeneity documented in Table~\ref{tab:noniid}.

\paragraph{FPGA projection methodology.}
Resource estimates target the Xilinx Zynq UltraScale+ ZCU102
(\texttt{xczu9eg-ffvb1156-2-e}): 274\,080~LUT, 2\,520~DSP,
912~BRAM36.
Projections follow the hls4ml scaling model \citep{hls4ml2021}:
$\text{LUT} = |\theta| \cdot b / 6$,
$\text{DSP} = |\theta| \cdot b / 18$,
$\text{Latency} = b/2~\mu$s at 500~MHz.
These are analytical projections; the FPGA estimation script is
available in the public repository.

\section{Results and Discussion}
\label{sec:results}

\begin{table}[htbp]
\centering
\caption{Test-set results: Mean\,$\pm$\,Std over 10 random seeds.
  $p$-values from two-tailed paired $t$-test vs.\ FP32
  ($n=10$, $\text{df}=9$). \textbf{Bold}: $p<0.05$.
  NASA score $S$ reported as sum over all test windows (Eq.~\ref{eq:nasa}).
  CV$_S$: coefficient of variation of $S$ across seeds.
  Cohen's $\hat{d}$ at $N=10$ carries 95\%~CI\,$\approx\hat{d}\pm0.95$;
  interpret as directional only.}
\label{tab:results}
\begin{tabular}{llccccc}
\toprule
\textbf{Sub.} & \textbf{Cfg} &
  \textbf{MAE (cycles)} & $p_{\text{MAE}}$ &
  \textbf{Score}~$S$ ($\times 10^{3}$) & $p_{S}$ & \textbf{CV}$_S$ \\
\midrule
\multirow{4}{*}{FD001}
  & FP32 & $17.52\pm0.47$ & --- & $449\pm123$ & --- & 27.3\% \\
  & INT8 & $17.51\pm0.48$ & 0.520 & $447\pm127$ & 0.746 & 28.6\% \\
  & INT4 & $17.48\pm0.51$ & 0.341 & $452\pm115$ & 0.802 & 25.3\% \\
  & INT2 & $19.03\pm1.62$ & \textbf{0.018}
         & $802\pm573$   & 0.064 & 72.0\% \\
\midrule
\multirow{4}{*}{FD002}
  & FP32 & $26.99\pm1.69$ & --- & $923\pm206$ & --- & 22.3\% \\
  & INT8 & $27.24\pm1.40$ & 0.265 & $951\pm167$ & 0.364 & 16.9\% \\
  & INT4 & $27.20\pm1.84$ & 0.264 & $938\pm233$ & 0.534 & 24.8\% \\
  & INT2 & $21.53\pm2.31$ & \textbf{0.001}$^\dagger$
         & $749\pm347$   & 0.207 & 45.8\% \\
\bottomrule
\multicolumn{7}{l}{$^\dagger$INT2 lower MAE on FD002 is an over-regularization artefact;} \\
\multicolumn{7}{l}{\phantom{$^\dagger$}see Section~\ref{sec:results_int2}.} \\
\end{tabular}
\end{table}

\subsection{INT8 Matches FP32 Across All Conditions}

INT8 achieves accuracy statistically indistinguishable from FP32
on both subsets and both metrics ($p \geq 0.265$ on all four
comparisons, Table~\ref{tab:results}).
This confirms the well-established result that 8-bit quantization
preserves model quality with negligible accuracy cost, consistent
with prior work \citep{alistarh2017}.

\begin{figure}[htbp]
  \centering
  \includegraphics[width=0.6\textwidth]{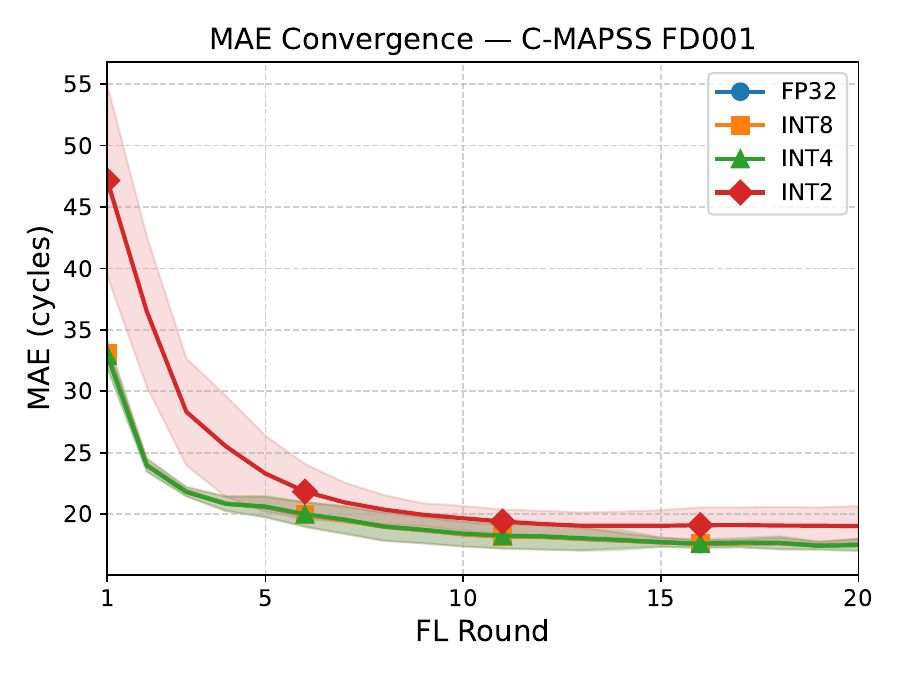}
  \caption{MAE convergence over 20 FL rounds on C-MAPSS FD001.
    Shaded bands: $\pm$1 std over 10 seeds.
    FP32, INT8, and INT4 converge to indistinguishable final MAE;
    INT2 exhibits slower convergence and higher variance.}
  \label{fig:mae}
\end{figure}

\subsection{INT4: Communication--Accuracy Parity}
\label{sec:results_int4}

The corrected multi-seed evaluation reveals no statistically significant
accuracy difference between INT4 and FP32 on either subset
($p>0.05$ on all metric$\times$subset combinations,
Table~\ref{tab:results}).
On FD001, the mean MAE difference is only 0.04~cycles,
well within the seed-to-seed variability of FP32 itself
(std\,=\,0.47~cycles).
On FD002, INT4 yields $p=0.264$ on MAE and $p=0.534$ on NASA score,
confirming full accuracy parity under the harder multi-condition
Non-IID setting.

\paragraph{LoRaWAN feasibility.}
INT4 delivers an $8\times$ reduction in gradient communication cost
(37.88~KiB $\to$ 4.73~KiB per round).
At 5~kbps, the 4.73~KiB INT4 payload requires $\approx 7.5$~s per
round; under a 1\% EU ISM-band duty-cycle limit, the minimum
inter-round interval is $\approx 12.5$~min, consistent with
predictive maintenance FL schedules where rounds are typically
spaced minutes to hours apart.

\paragraph{Main claim.}
\emph{INT4 maintains accuracy statistically indistinguishable from
FP32 ($p > 0.05$ on all comparisons, both subsets) while delivering
$8\times$ communication reduction, making it the practical operating
point for bandwidth-constrained aerospace IoT deployments.}

\subsection{Methodological Bias of IID Partitioning}
\label{sec:iid_bias}

Table~\ref{tab:iid_bias} compares FP32 and INT4 on FD001 under both
partitioning strategies, evaluated over 10 seeds.
Under the artificial IID partition, the NASA score variance is
suppressed (std\,=\,31k vs.\ 123k under Non-IID), and INT4 can appear
to marginally outperform FP32.
Under the realistic Non-IID partition, the true cross-seed variance is
revealed, correctly establishing statistical parity rather than dominance.

This finding has a broader implication: evaluation protocols that
assign training data to clients by random index shuffling (IID) rather
than by engine assignment (Non-IID) will systematically underestimate
prediction variance and may incorrectly conclude that quantization
provides an accuracy benefit, when in reality it does not.

\begin{table}[htbp]
\centering
\caption{Methodological bias: IID vs.\ Non-IID partitioning on FD001
  (mean\,$\pm$\,std over 10 seeds).
  IID suppresses variance, making INT4 appear to outperform FP32;
  Non-IID reveals statistical parity.}
\label{tab:iid_bias}
\begin{tabular}{llcc}
\toprule
\textbf{Partition} & \textbf{Config} &
  \textbf{MAE (cycles)} & \textbf{Score}~$S$ ($\times 10^{3}$) \\
\midrule
\multirow{2}{*}{IID}
  & FP32 & $17.34\pm0.36$ & $421\pm41$ \\
  & INT4 & $17.28\pm0.26$ & $440\pm31$ \\
\midrule
\multirow{2}{*}{Non-IID}
  & FP32 & $17.52\pm0.47$ & $449\pm123$ \\
  & INT4 & $17.48\pm0.51$ & $452\pm115$ \\
\bottomrule
\end{tabular}
\end{table}

Whether gradient quantization acts as a genuine implicit regularizer
under Non-IID FL \citep{he2025,ma2022} remains an open question;
the evidence presented here does not confirm this hypothesis at
$\alpha=0.05$ on either subset.

\subsection{INT2: Instability and Non-Reproducibility}
\label{sec:results_int2}

\begin{figure}[htbp]
  \centering
  \includegraphics[width=0.9\textwidth]{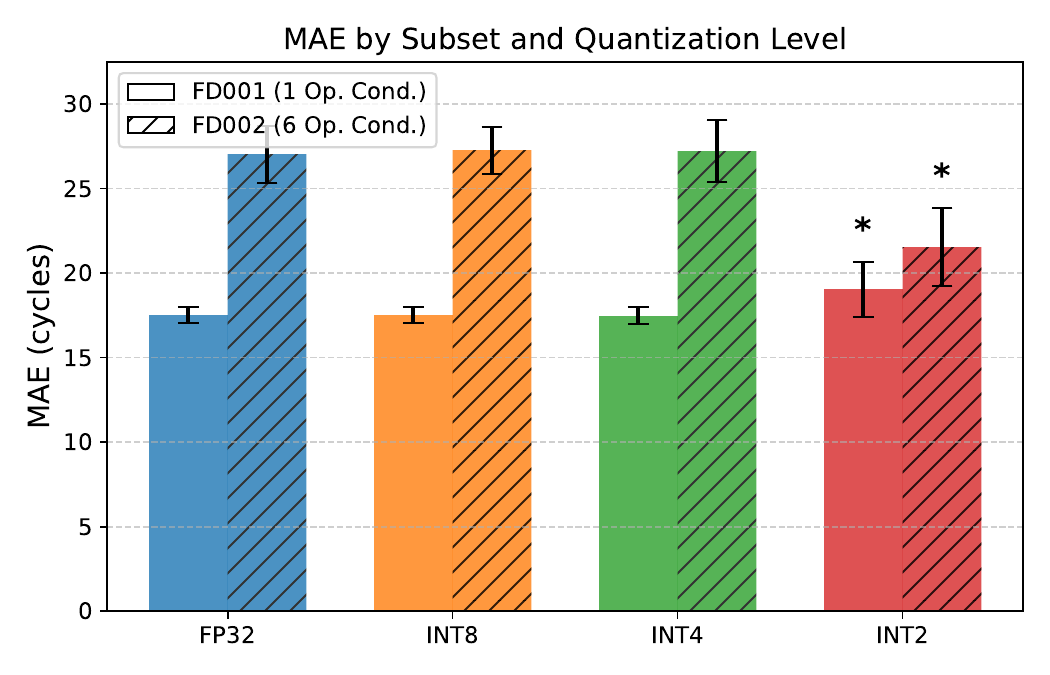}
  \caption{MAE by subset and quantization level. Hatching indicates
    FD002 (6 operating conditions). Asterisk (*): $p < 0.05$ vs.\
    FP32 on MAE. \textbf{Warning:} The lower MAE of INT2 on FD002 is
    an over-regularization artefact, not a genuine accuracy improvement
    (see Section~\ref{sec:results_int2}).}
  \label{fig:fd_comparison}
\end{figure}

INT2 behaviour differs qualitatively between the two subsets and
cannot be characterised as uniformly degrading or uniformly beneficial.
Unlike classification settings where binary or 1-bit neural networks
can achieve competitive accuracy \citep{lee2025}, INT2 proves
fundamentally unsuitable for safety-critical RUL regression.

\paragraph{FD001 (single operating condition).}
INT2 MAE is significantly worse than FP32
($19.03\pm1.62$ vs.\ $17.52\pm0.47$ cycles, $+8.6\%$, $p=0.018$).
The NASA score is not significantly different from FP32 ($p=0.064$),
but the coefficient of variation is 72.0\% compared to 27.3\% for
FP32, indicating severe seed-to-seed instability.

\paragraph{FD002 (six operating conditions).}
INT2 achieves a statistically significant \emph{lower} MAE than FP32
($21.53\pm2.31$ vs.\ $26.99\pm1.69$ cycles, $-20.2\%$, $p=0.001$).
This apparent improvement is, however, an over-regularization artefact:
the extreme precision constraint of INT2 forces weight updates onto the
3-level grid $\{-\alpha^{(l)}, 0, +\alpha^{(l)}\}$---effectively acting
as a per-layer ternary update with a dynamic scale rather than a standard
uniform 2-bit grid---preventing the model from adapting to the
heterogeneous six-condition distribution of FD002 in the same way as
higher-precision configurations.
The result is a form of underfitting that accidentally achieves lower
MAE on some seeds by predicting conservatively, not by genuinely
learning the degradation pattern.

The NASA score confirms this diagnosis: the mean score for INT2 is
$749{,}000 \pm 347{,}000$ (CV\,=\,45.8\%) versus
$923{,}000 \pm 206{,}000$ (CV\,=\,22.3\%) for FP32.
While the mean score is lower for INT2, the variance is
$\approx 2\times$ higher, and individual seeds produce wildly
divergent outcomes.
In a safety-critical predictive maintenance context, a model with
CV\,=\,45.8\% on the NASA asymmetric score is operationally unusable
regardless of its average MAE. This dynamic is visually summarized
in Figure~\ref{fig:fd_comparison}.

\begin{figure}[htbp]
  \centering
  \includegraphics[width=0.6\textwidth]{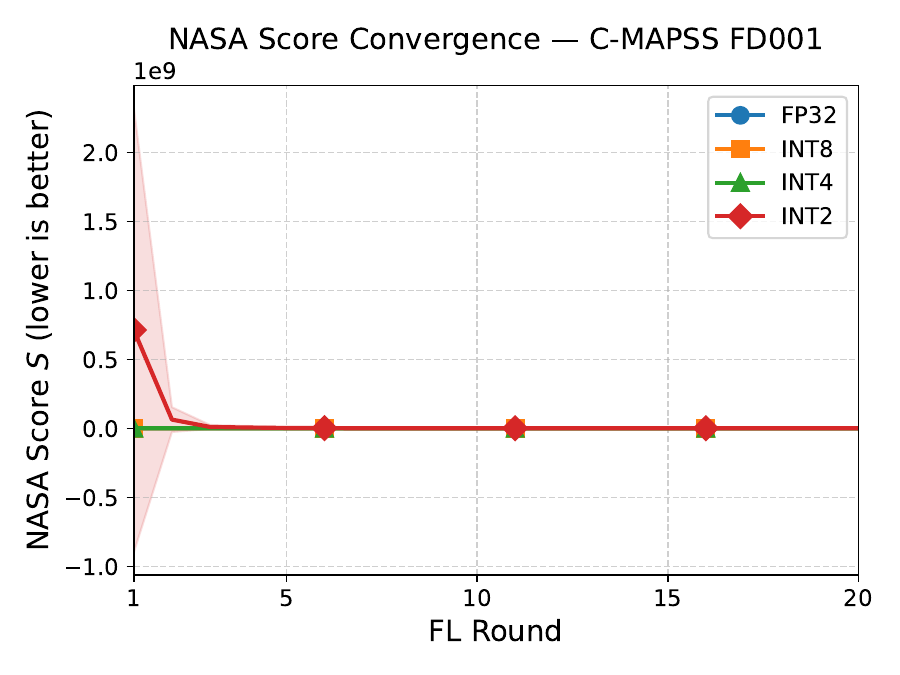}
  \caption{NASA score $S$ convergence on C-MAPSS FD001. Lower is better.
    INT2 early-round values reach $10^9$ and are off-scale; the y-axis
    is clipped for readability. Negative values arise when systematic
    under-prediction dominates; INT2 oscillates between extreme positive
    and negative scores, illustrating non-reproducibility.}
  \label{fig:score}
\end{figure}

\paragraph{Verdict.}
INT2 is unsuitable for aerospace RUL regression not because of
uniform accuracy degradation, but because of \emph{fundamental
non-reproducibility}: the interaction between the 3-level quantization
grid and Non-IID operating conditions produces outcomes that vary
catastrophically across initializations, precluding reliable deployment.

\subsection{Accuracy--Communication Trade-off}

\begin{figure}[htbp]
  \centering
  \includegraphics[width=0.6\textwidth]{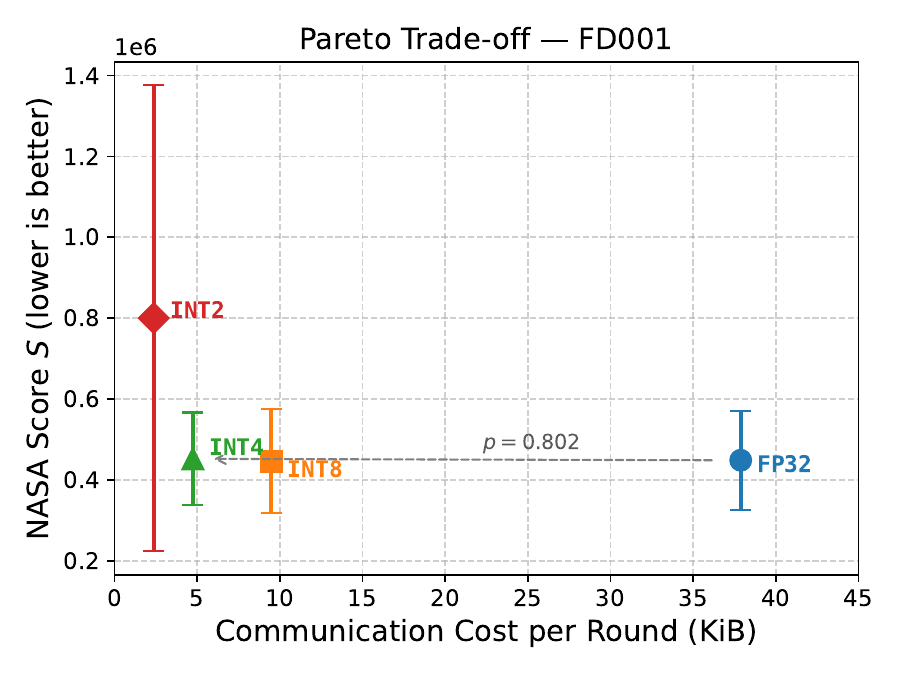}
  \caption{Accuracy--communication trade-off on FD001.
    Error bars: $\pm$1 std over 10 seeds.
    Arrow: FP32 $\to$ INT4 operating point.
    Annotated $p$-value from paired $t$-test on NASA score
    (INT4 vs.\ FP32).}
  \label{fig:pareto}
\end{figure}

Figure~\ref{fig:pareto} plots the accuracy--communication Pareto front
on FD001. The FP32 $\to$ INT4 path achieves an $8\times$ communication
reduction with $p=0.802$ on NASA score, confirming that the gain is
not statistically distinguishable from the baseline.
INT8 offers a 4$\times$ reduction at $p=0.746$.
INT2 achieves the lowest communication cost but is excluded from
the Pareto front due to its instability.

\subsection{FPGA Feasibility}
\label{sec:fpga}

Table~\ref{tab:fpga} lists the analytical FPGA resource projections.
The DSP count is the binding resource constraint for all configurations.
FP32 requires 684\% of available DSPs; INT8 requires 171\%.
Only INT4 and INT2 fit the ZCU102, with INT4 at 85.5\% DSP utilisation
and INT2 at 42.7\%.

INT4 leaves 366 spare DSPs, which could potentially host an NTT-based
homomorphic encryption co-processor \citep{nguyen2023}, enabling
encrypted gradient transmission at $2~\mu$s inference latency.
The ARM Cortex-A53 on the ZCU102 PS quad-core would execute FL local
training and INT4 quantization in software (PyTorch AArch64), while
the PL fabric accelerates INT4 inference via hls4ml, potentially
enabling a complete training--quantization--inference pipeline on a
single SoC.

\emph{All figures in Table~\ref{tab:fpga} are analytical projections
derived from the hls4ml scaling model and have not been validated
against physical ZCU102 synthesis reports.}

\begin{table}[htbp]
\centering
\caption{Analytical FPGA resource projections --- Xilinx ZCU102
  (274\,080~LUT $|$ 2\,520~DSP $|$ 912~BRAM36).
  Latency at 500~MHz, sequence length 50.
  Comm.\ cost excludes per-layer scale overhead
  ($8\times4$~B\,$\approx$\,0.03~KiB).
  Projections derived from hls4ml scaling model;
  not validated on physical silicon.}
\label{tab:fpga}
\begin{tabular}{lrrrrrr}
\toprule
\textbf{Cfg} &
  \textbf{LUT} & \textbf{\%LUT} &
  \textbf{DSP} & \textbf{\%DSP} &
  \textbf{Lat.} & \textbf{Fit} \\
\midrule
FP32 & 51\,717 & 18.9\% & 17\,239 & 684.1\% & 16~$\mu$s & $\times$ \\
INT8 & 12\,929 &  4.7\% &  4\,309 & 171.0\% &  4~$\mu$s & $\times$ \\
INT4 &  6\,464 &  2.4\% &  2\,154 &  85.5\% &  2~$\mu$s & $\checkmark$ \\
INT2 &  3\,232 &  1.2\% &  1\,077 &  42.7\% &  1~$\mu$s & $\checkmark$ \\
\bottomrule
\end{tabular}
\end{table}

\section{Conclusion}
\label{sec:conclusion}

This paper investigated gradient quantization in a federated learning
system for aerospace predictive maintenance on the NASA C-MAPSS benchmark.

The primary methodological contribution is demonstrating that na\"ive
IID client partitioning artificially inflates the apparent accuracy
benefit of quantization.
Under correct Non-IID evaluation with ground-truth test RUL labels and
a proper sliding-window test protocol, INT4 achieves accuracy parity
with FP32 ($p>0.05$ on all metric$\times$subset combinations) while
delivering an $8\times$ communication reduction, making it the practical
operating point for bandwidth-constrained aerospace IoT deployments.

INT2 exhibits qualitatively different behaviour across subsets: MAE
degrades significantly on FD001 ($+8.6\%$, $p=0.018$), while an
apparent MAE improvement on FD002 ($-20.2\%$, $p=0.001$) is identified
as an over-regularization artefact.
In both cases, INT2 is rendered operationally unusable by catastrophic
NASA score instability (CV\,=\,72.0\% on FD001, 45.8\% on FD002),
confirming non-reproducibility under heterogeneous operating conditions.

Analytical FPGA projections show that INT4 fits within the Xilinx
ZCU102 at 85.5\% DSP utilisation, leaving 366 spare DSPs for potential
cryptographic co-design, subject to silicon validation.

Future work will (i)~incorporate a formal $(\varepsilon,\delta)$-DP
analysis against gradient-inversion threat models
\citep{zhu2019,geiping2020}, (ii)~validate FPGA projections on physical
ZCU102 silicon, (iii)~quantify Non-IID severity via EMD across a
broader engine-partitioning parameter sweep, and (iv)~extend the
evaluation to additional C-MAPSS subsets (FD003, FD004) and to
federated settings with heterogeneous client hardware.

\section*{Data and Code Availability}
The PyTorch implementation of AeroConv1D, the full federated learning
simulation framework, raw experimental logs (10-seed Non-IID and IID
partitions), and FPGA estimation scripts are openly available at:
\begin{center}
\url{https://github.com/therealdeadbeef/aerospace-fl-quantization}
\end{center}
The NASA C-MAPSS dataset is publicly available via the NASA Prognostics
Data Repository \citep{saxena2008}.

\bibliographystyle{abbrvnat}
\bibliography{references}

@article{saxena2008,
author = {Saxena, Abhinav and Goebel, Kai and Simon, Don and Eklund, Neil},
year = {2008},
month = {10},
pages = {},
title = {Damage propagation modeling for aircraft engine run-to-failure simulation},
journal = {International Conference on Prognostics and Health Management},
doi = {10.1109/PHM.2008.4711414}
}

@InProceedings{mcmahan2017,
  title = 	 {{Communication-Efficient Learning of Deep Networks from Decentralized Data}},
  author = 	 {McMahan, Brendan and Moore, Eider and Ramage, Daniel and Hampson, Seth and Arcas, Blaise Aguera y},
  booktitle = 	 {Proceedings of the 20th International Conference on Artificial Intelligence and Statistics},
  pages = 	 {1273--1282},
  year = 	 {2017},
  editor = 	 {Singh, Aarti and Zhu, Jerry},
  volume = 	 {54},
  series = 	 {Proceedings of Machine Learning Research},
  month = 	 {20--22 Apr},
  publisher =    {PMLR},
  url = 	 {https://proceedings.mlr.press/v54/mcmahan17a.html},
}

@inproceedings{alistarh2017,
 author = {Alistarh, Dan and Grubic, Demjan and Li, Jerry and Tomioka, Ryota and Vojnovic, Milan},
 booktitle = {Advances in Neural Information Processing Systems},
 editor = {I. Guyon and U. Von Luxburg and S. Bengio and H. Wallach and R. Fergus and S. Vishwanathan and R. Garnett},
 pages = {},
 publisher = {Curran Associates, Inc.},
 title = {QSGD: Communication-Efficient SGD via Gradient Quantization and Encoding},
 url = {https://proceedings.neurips.cc/paper_files/paper/2017/file/6c340f25839e6acdc73414517203f5f0-Paper.pdf},
 volume = {30},
 year = {2017}
}

@inproceedings{bernstein2018,
  title={signSGD: compressed optimisation for non-convex problems},
  author={Jeremy Bernstein and Yu-Xiang Wang and Kamyar Azizzadenesheli and Anima Anandkumar},
  booktitle={International Conference on Machine Learning},
  year={2018},
  url={https://api.semanticscholar.org/CorpusID:7763588}
}

@article{ma2022,
author = {Ma, Xiaodong and Zhu, Jia and Lin, Zhihao and Qin, Yangjie and Chen, Shanxuan},
year = {2022},
month = {05},
pages = {},
title = {A state-of-the-art survey on solving non-IID data in Federated Learning},
volume = {135},
journal = {Future Generation Computer Systems},
doi = {10.1016/j.future.2022.05.003}
}

@inproceedings{purkayastha2024,
  title     = {Federated Learning for Predictive Maintenance: A Survey of Methods, Applications, and Challenges},
  author    = {Purkayastha, A. A. and others},
  booktitle = {2024 IEEE 67th International Midwest Symposium on Circuits and Systems (MWSCAS)},
  year      = {2024},
  organization = {IEEE}
}

@article{he2025,
  title     = {FedDT: A Communication-Efficient Federated Learning via Knowledge Distillation and Ternary Compression},
  author    = {He, Z. and others},
  journal   = {Electronics},
  volume    = {14},
  number    = {11},
  pages     = {2183},
  year      = {2025},
  publisher = {MDPI}
}

@article{hls4ml2021,
  title     = {hls4ml: An Open-Source Codesign Workflow to Empower Scientific Low-Power Machine Learning Devices},
  author    = {Fahim, Farah and others},
  journal   = {IEEE Transactions on Nuclear Science},
  volume    = {68},
  number    = {8},
  pages     = {1885--1896},
  year      = {2021},
  publisher = {IEEE}
}

@inproceedings{nguyen2023,
  title     = {CKKS-Based Homomorphic Encryption Architecture Using Parallel NTT Multiplier},
  author    = {Nguyen, Tran Duc Duy and Kim, J. and Lee, H.},
  booktitle = {2023 IEEE International Symposium on Circuits and Systems (ISCAS)},
  year      = {2023},
  organization = {IEEE}
}

@article{ye2025,
  title     = {Implementing Homomorphic Encryption-Based Logic Locking in SoC Designs},
  author    = {Ye, Zheng and Ikeda, Makoto},
  journal   = {IEEE Transactions on Very Large Scale Integration (VLSI) Systems},
  volume    = {33},
  number    = {7},
  year      = {2025},
  publisher = {IEEE}
}

@article{laouiti2025,
  title     = {Hardware Acceleration of Fully Homomorphic Encryption for Edge Federated Learning},
  author    = {Laouiti, A. and others},
  journal   = {IEEE Internet of Things Journal},
  year      = {2025},
  publisher = {IEEE}
}

@inproceedings{zhu2019,
  title     = {Deep Leakage from Gradients},
  author    = {Zhu, Ligeng and Liu, Zhijian and Han, Song},
  booktitle = {Advances in Neural Information Processing Systems (NeurIPS)},
  volume    = {32},
  year      = {2019}
}

@inproceedings{geiping2020,
  title     = {Inverting Gradients -- How Easy Is It to Break Privacy in Federated Learning?},
  author    = {Geiping, Jonas and Bau, Hendrik and Droste, Finn and Moeller, Michael},
  booktitle = {Advances in Neural Information Processing Systems (NeurIPS)},
  volume    = {33},
  pages     = {16937--16947},
  year      = {2020}
}

@article{khalil2023,
  title     = {A Federated Learning Model Based on Hardware Acceleration for the Early Detection of Alzheimer's Disease},
  author    = {Khalil, K. and others},
  journal   = {Sensors},
  volume    = {23},
  number    = {19},
  pages     = {8272},
  year      = {2023},
  publisher = {MDPI}
}

@inproceedings{wang2023,
  title     = {SAM: A Scalable Accelerator for Number Theoretic Transform Using Multi-Dimensional Decomposition},
  author    = {Wang, C. and Gao, M.},
  booktitle = {Proceedings of the IEEE/ACM International Conference on Computer-Aided Design (ICCAD)},
  year      = {2023}
}

@misc{landau2025,
      title={Federated learning framework for collaborative remaining useful life prognostics: an aircraft engine case study}, 
      author={Diogo Landau and Ingeborg de Pater and Mihaela Mitici and Nishant Saurabh},
      year={2025},
      eprint={2506.00499},
      archivePrefix={arXiv},
      primaryClass={cs.LG},
      url={https://arxiv.org/abs/2506.00499}, 
}

@article{lee2025,
  title     = {BiPruneFL: Computation and Communication Efficient Federated Learning With Binary Quantization and Pruning},
  author    = {Lee, Sangmin and others},
  journal   = {IEEE Access},
  year      = {2025},
  publisher = {IEEE},
}

@book{hedges1985,
  title     = {Statistical Methods for Meta-Analysis},
  author    = {Hedges, Larry V. and Olkin, Ingram},
  year      = {1985},
  publisher = {Academic Press}
}

@misc{li2016,
      title={Ternary Weight Networks}, 
      author={Fengfu Li and Bin Liu and Xiaoxing Wang and Bo Zhang and Junchi Yan},
      year={2022},
      eprint={1605.04711},
      archivePrefix={arXiv},
      primaryClass={cs.CV},
      url={https://arxiv.org/abs/1605.04711}, 
}

@misc{zheng2025fedhqhybridruntimequantization,
      title={FedHQ: Hybrid Runtime Quantization for Federated Learning}, 
      author={Zihao Zheng and Ziyao Wang and Xiuping Cui and Maoliang Li and Jiayu Chen and Yun and Liang and Ang Li and Xiang Chen},
      year={2025},
      eprint={2505.11982},
      archivePrefix={arXiv},
      primaryClass={cs.LG},
      url={https://arxiv.org/abs/2505.11982}, 
}

\end{document}